\documentclass{egpubl}
\usepackage{pg2025s}

\WsConferencePaper

\usepackage[T1]{fontenc}
\usepackage{dfadobe}  

\usepackage{cite}  
\BibtexOrBiblatex
\electronicVersion
\PrintedOrElectronic
\ifpdf \usepackage[pdftex]{graphicx} \pdfcompresslevel=9
\else \usepackage[dvips]{graphicx} \fi

\usepackage{egweblnk} 

\usepackage{amssymb}
\usepackage{amsmath}
\usepackage{algpseudocode}
\usepackage{algorithm}
\usepackage{multirow}
\newtheorem{theorem}{Theorem}
\usepackage{hyperref}
\usepackage{cleveref}

\usepackage{pgfplots} 
\usepackage{subcaption}

\usepackage{booktabs}
\usepackage{xcolor}
\usepackage{makecell}

\title[VF-Plan: Bridging the Art Gallery Problem and Static LiDAR Scanning with Visibility Field Optimization]%
      {VF-Plan: Bridging the Art Gallery Problem and Static LiDAR Scanning with Visibility Field Optimization}

\author[B. Xiong et al.]
{\parbox{\textwidth}{\centering B. Xiong$^{1}$\thanks{Corresponding author: b.xiong@whut.edu.cn}\orcid{0000-0001-7756-0901},
L. Zhang$^{1}$, 
R. Huang$^{1}$, 
J. Zhou$^{1}$\orcid{0000-0002-6094-1203}, 
S.R.U.N. Jafri$^{2}$\orcid{0000-0002-2097-828X}, 
B. Wu$^{3}$\thanks{Corresponding author: ustcbjwu@gmail.com}\orcid{0009-0007-1945-8707}, 
F. Li$^{4}$\orcid{0000-0001-9443-777X}\\
$^1$ Wuhan University of Technology,
        $^2$ NED University of Engineering and Technology\\
        $^3$ Zhejiang University,
        $^4$ The Advanced Laser Technology Laboratory of Anhui Province
    }
}

\begin{document}

\teaser{
 \includegraphics[width=0.75\linewidth]{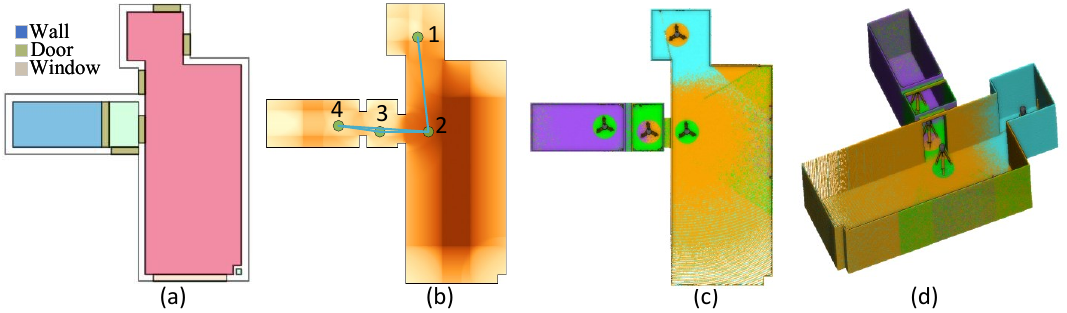}
 \centering
 \caption{Viewpoint optimization for static LiDAR Scanning. (a) The input floor plan; (b) Optimized viewpoints indicated by green dots, with the Visibility Field displayed as the background image; (c) and (d) Simulated LiDAR points from optimal viewpoints, shown in 2D and 3D respectively, with each scan distinguished by a unique color.}
 \label{fig:Teaser.}
}

\maketitle
\begin{abstract}
Viewpoint planning is critical for efficient 3D data acquisition in applications such as 3D reconstruction, building life-cycle management, navigation, and interior decoration. However, existing methods often neglect key optimization objectives specific to static LiDAR systems, resulting in redundant or disconnected viewpoint networks. The viewpoint planning problem (\textbf{VPP}) extends the classical Art Gallery Problem (\textbf{AGP}) by requiring full coverage, strong registrability, and coherent network connectivity under constrained sensor capabilities. To address these challenges, we introduce a novel Visibility Field (\textbf{VF}) that accurately captures the directional and range-dependent visibility properties of static LiDAR scanners. We further observe that visibility information naturally converges onto a 1D skeleton embedded in the 2D space, enabling significant searching space reduction. Leveraging these insights, we develop a greedy optimization algorithm tailored to the VPP, which constructs a minimal yet fully connected Viewpoint Network (\textbf{VPN}) with low redundancy. Experimental evaluations across diverse indoor and outdoor scenarios confirm the scalability and robustness of our method. Compared to expert-designed VPNs and existing state-of-the-art approaches, our algorithm achieves comparable or fewer viewpoints while significantly enhancing connectivity. In particular, it reduces the weighted average path length by approximately 95\%, demonstrating substantial improvements in compactness and structural efficiency. Code is available at \url{https://github.com/xiongbiaostar/VFPlan}.


\end{abstract}  
\section{Introduction}\label{sec:intro}
The VPP has emerged as a critical research spot due to the growing use of LiDAR and camera sensors in applications such as 3D reconstruction, autonomous navigation, facility assessment, building life-cycle management, and interior decoration applications~\cite{scott2003view,ma2023review}. VPP focuses on determining the optimal placements and configurations of sensors to achieve specific reconstruction goals, such as maximizing observation coverage and minimizing occlusions, while balancing efficiency and data quality. Both model-free and model-based algorithms for viewpoint and path planning have gained significant attention due to their applicability across various industries, including robotics, construction, and surveillance~\cite{maboudi2023}.

Conceptually, VPP is rooted in the AGP, a classic challenge in computational geometry that seeks the minimal number of “guards” required to cover a given area, typically a polygon, with full 360° visibility~\cite{lee1986}. AGP is often simplified by constraining guard placements to polygon vertices; however, this simplification does not capture the true computational complexity of the problem. In its generalized form, AGP is $\exists \mathbb{R}$-complete, meaning it is at least as hard as any NP-complete problem, with no known efficient solutions~\cite{couto2011, Abrahamsen2021}. This NP-hard nature has inspired substantial research interest, particularly in constrained versions that make idealized assumptions, such as complete visibility and static scenes~\cite{couto2011, scott2003view}.


VPP, however, introduces additional complexities beyond those of AGP. Unlike static guards, sensors in VPP must maintain overlapping fields of view for accurate data registration, accommodate limited and directional visibility ranges, and meet high accuracy requirements for 3D scene reconstruction. These real-world demands, especially in cluttered or partially obstructed environments, render VPP a more challenging NP-hard problem, requiring innovative optimization strategies beyond those used for AGP~\cite{scott2003view, Abrahamsen2021}.

We address the challenges of static LiDAR coverage planning by introducing a novel greedy algorithm based on the concept of a \textit{Visibility Field (VF)}. The VF encapsulates the physical constraints of LiDAR sensors, including limited sensing range and angle-dependent visibility, enabling a substantial reduction in computational complexity. We formally demonstrate that visibility information is inherently concentrated along a one-dimensional skeleton embedded within the two-dimensional space. This observation allows us to restrict the search space from 2D to 1D by targeting critical structural features such as the medial axis and junctions within the VF. Consequently, we construct a minimal, fully connected network of viewpoints that achieves complete scene coverage with reduced redundancy. Our main contributions are summarized as follows:

\begin{enumerate}
    \item \textbf{Unified Solution to VPP and AGP:} Our method addresses the combined challenges of viewpoint planning and the AGP, achieving efficient coverage with minimal viewpoints and robust network connectivity.
    
    \item \textbf{Visibility Field for Dimensionality Reduction:} We introduce a continuousVF tailored to static LiDAR systems, reducing the optimization space from 2D to 1D by leveraging structural elements like medial axis and joints.
    
    \item \textbf{Greedy Algorithm for Efficient Coverage:} Our greedy algorithm utilizes the VF’s structural insights to construct a minimal, fully connected VPN, ensuring both robust coverage and network connectivity.
\end{enumerate}
\begin{figure*}[tp]
    \centering
    \includegraphics[width=0.85\textwidth]{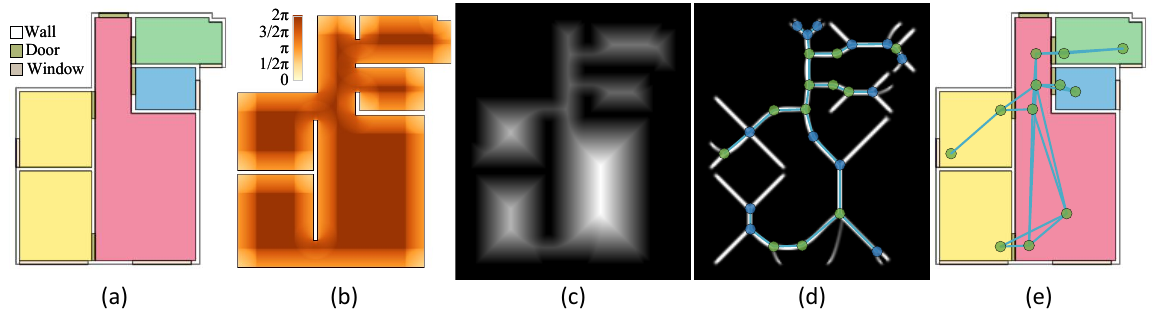}
    \caption{Pipeline of viewpoint network (VPN) optimization using VF. (a) Input floor plan with walls, doors, and windows; (b) Computed VF encoding the valid observed angle at each pixel; (c) Distance field derived from the VF; 
    (d) Extraction of converging lines (medial axis) and converging points (joints) used as candidate viewpoints; (e) Final optimized VPN ensuring full coverage and strong connectivity.}
    \label{fig:pipeline}
\end{figure*}

\section{Related Work}

This section reviews key advancements in sensor network planning, covering both static and mobile sensing, with a focus on recent works in LiDAR and computer vision area. These serve as the foundation for the viewpoint planning method proposed here, addressing gaps in efficient coverage, connectivity, and optimization complexity.

\subsection{Station-Based Scanning}
Sensor network design shares similarities with the Planning for Scanning (P4S) problem, observed in applications like surveillance cameras~\cite{Morsly2011,Zhao2013,Ghanem2015}, directional sensors~\cite{Fusco2009}, 5G base stations~\cite{Shi2020,Wang2020}, stealth game~\cite{xu2014generative}, and station-based LiDAR scanning~\cite{aryan2021}. For LiDAR, the goal is to cover the entire scene with comprehensive wall scanning while ensuring data quality criteria such as completeness and accuracy~\cite{liu2021,aryan2021}.

Advancements in LiDAR technology have led to P4S methods using prior information like 2D drawings and 3D models~\cite{liu2021}. These methods have been applied to civil infrastructures~\cite{ xu2025two}, space frame structures~\cite{liu2021}, rebar detection~\cite{Li2021,li2025full}, and landslide monitoring~\cite{zhang2023}. In cases where prior information is outdated or unavailable, low-resolution preliminary scans are required, either indoors~\cite{qiu2021} or outdoors using UAVs~\cite{Li2022}. 

P4S optimizations aim to maximize coverage and overlap while minimizing the number of viewpoints~\cite{Ibrahim2022, chen2025automated}. Methods like CMA Evolution Strategy~\cite{Rougeron2022} and regular grid sampling~\cite{Jia2019,Noichl2022} increase the likelihood of finding optimal viewpoints but also raise computational complexity. Jia's approach using grid-based sampling~\cite{Jia2018,Jia2019,Jia2022} scores candidate viewpoints based on observed wall segments, though this approach is limited by its reliance on reference objects for registration. Noichl~\cite{noichl2024automating} improve Jia's method into 3D scene for an uniform covering by triangulating 3D scene into meshes and checking their visibility. For scenarios lacking BIM priors, structural elements can be surveyed and detected using low-cost sensors to serve as a preliminary prior for static station planning. A precise scan is then performed in a stop-and-go manner using a static scanner mounted on a robotic dog~\cite{chung2025automated, hu2025semantic}.

Our approach addresses these limitations by using a continuous VF model, which reduces optimization from 2D to 1D by focusing on structural convergence points like medial axis and joints, thus minimizing computation and enhancing adaptability.

\subsection{Mobile-Based Exploration}
Coverage Path Planning (CPP) focuses on computing optimal, collision-free paths for robots to fully cover target areas~\cite{Tan2021}. Techniques like Rapidly-exploring Random Trees (RRT) and their variants (RRT*~\cite{Karaman2011}, RRT\#~\cite{Otte2015}) are widely used for autonomous exploration in unknown spaces, including UAVs~\cite{maboudi2023} and wheeled robots~\cite{Placed2023}. RRT-based methods maximize information gain while minimizing travel distance~\cite{Hepp2018,Xie2018,Zhang2021}.

Methods utilizing tensor fields for guiding robot movements enable efficient path routing in partially reconstructed indoor scenes~\cite{Xu2017, Xi2024}. Recent approaches like the Signed Distance Field (SDF) and Hamilton-Jacobi skeleton further improve path planning in known environments~\cite{Noel2023}. These concepts share similarities with our VF, which leverages converging lines for efficient viewpoint planning.

For known environments, algorithms like the receding horizon "next-best-view" approach~\cite{Bircher2016} and dual-resolution mapping schemes~\cite{Cao2023} balance detailed local mapping with efficient global exploration. For applications in autonomous driving and exploration, recent advances such as "LookOut"~\cite{cui2021lookout} propose diverse multi-future prediction models that enhance autonomous navigation by predicting possible trajectories, balancing safety, and efficiency.


\section{Problem Definition and Global Optimality}\label{sec:theory}

We formalize the VPP for static LiDAR scanning across a collection of two-dimensional polygonal regions \( \mathcal{P} = \{P_1, P_2, \dots, P_k\} \), each comprising structural boundaries and possible internal obstacles. The boundaries of these regions are uniformly discretized into a set of equal-length line segments \( L = \{l_1, l_2, \dots, l_n\} \), representing the structural elements to be scanned. The objective is to select a set of viewpoints \( V = \{v_1, v_2, \dots, v_m\} \) that satisfies two core constraints: \emph{coverage} and \emph{connectivity}.

\textbf{Coverage:} Every segment \( l_j \in L \) must be visible from at least one viewpoint \( v_i \in V \). We formalize this visibility relationship using a binary coverage table \( C \), where each entry \( C_{ij} \) indicates whether \( v_i \) observes \( l_j \):
\begin{equation}\label{eq:coverage}
    C_{ij} = \begin{cases} 
    1 & \text{if } v_i \text{ sees } l_j, \\ 
    0 & \text{otherwise}.
    \end{cases}
\end{equation}

\textbf{Connectivity:} Let the \emph{overlap ratio} $O_{ik}$ as the fraction of $L$ observed by both viewpoints $v_i$ and $v_k$. The induced subgraph $(S, E_s)$ is connected under the overlap threshold $\tau$, where
\begin{equation}\label{eq:connectivity}
    E_S \;=\;\bigl\{(v_i,v_k)\in E \;\big\vert\; v_i,v_k\in S,\;O_{ik}\ge\tau\bigl\}.
\end{equation}
We aim to find the smallest subset $S \subseteq V$ such that:
\begin{equation}\label{eq:problem_formulation}
\min_{S \subseteq V}\; |S| \quad \text{subject to: } (\ref{eq:coverage})~and~(\ref{eq:connectivity}) 
\end{equation}

This NP-hard problem extends the classical set cover problem by incorporating additional connectivity constraints. We demonstrate that restricting $V$ to the \emph{medial axis} (MA) preserves both complete coverage and network connectivity. As observation data in the VF naturally converge toward the MA and its joint points, placing viewpoints along these skeletal structures ensures comprehensive coverage. This property reduces VPN optimization from a 2D search to a 1D problem constrained to the MA. Consequently, a minimal cover on the MA remains globally optimal. Formal proofs of visibility completeness and global optimality for this reduction are provided in Appendix~\ref{appendix:theory}.

\section{Method} \label{sec:method}

\subsection{Overview}
Our proposed method, VF-Plan, addresses the NP-hard complexity of the View Planning Problem (VPP) through a greedy strategy centered on the VF. Given a structural 2D floorplan, VF-Plan constructs a VPN that ensures full coverage with a minimal number of viewpoints, while maintaining connectivity and registrability. The selected viewpoints form a fully connected observation network, which can subsequently be used to minimize registration errors via Least Squares Fitting (LSF).

As shown in \cref{fig:pipeline}, VF-Plan starts from a floorplan containing structural elements such as walls, columns, doors, and windows. The process includes VF computation, medial axis and joint-point extraction, and candidate viewpoint initialization. Overlap ratios between candidates are then evaluated, and a greedy algorithm selects the minimal subset forming a connected VPN. To accelerate the visibility checks for each viewpoint, we employ a BSP-tree~\cite{BSPTree} with balanced splitting and no depth limit, starting from the median line at each iteration. Prior to BSP-tree construction and VF computation, the floorplan is preprocessed into closed polygons while preserving all structural elements.

\subsection{Visibility Field (VF)}

Laser scanning is limited to visible surfaces, leaving occluded areas unobserved. Therefore, visibility and coverage analysis are essential for viewpoint planning, with key objectives of maximizing observation Coverage and registrability. The field of view is central to VPN optimization, defining the information captured at each viewpoint and guiding the selection of optimal viewpoints.

A common method for evaluating the information scanned by a LiDAR sensor is based on the length of observed walls \cite{Jia2022}. Typically, wall length is counted either by ensuring both endpoints are visible or by partitioning walls into uniformly sampled segments within the scanner’s range. For 3D cases, visibility is evaluated by counting the number of observed voxels. However, these approaches lack normalization, yielding values that vary with discretization resolution and do not account for differences in distance or orientation to the scanner, impacting computational stability and load without improving coverage accuracy.

\begin{figure}
    \centering
    \includegraphics[width=0.3\linewidth]{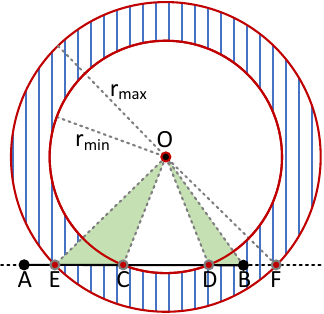}
    \caption{Scan model of a static scanner. The scanner provides valid observations within a 360° angle and a range from \( r_{\text{min}} \) to \( r_{\text{max}} \).}
    \label{fig:scanner_model}
\end{figure}

Our approach instead uses the valid scanned angle to integrate relevant factors into a single index, uniformly scaled from 0 to \(2\pi\). We define a static LiDAR scanner with a 360° field of view in the X-Y plane, a range from \(r_{\text{min}}\) to \(r_{\text{max}}\), and uniform scanning speed, as shown in \cref{fig:scanner_model}. Only objects within this range are observable. Generally, closer viewpoints collect more information, but excessive proximity can occlude other parts of the field. Scanning is performed in discrete observation angles, with observed information defined as the valid observed angle, ensuring alignment with scanner properties.

To quantify observed data, let Circle\_min intersect line segment \(AB\) at points \(C\) and \(D\), and Circle\_max at points \(E\) and \(F\). The valid observed line segment is:
\begin{equation}
L_{\text{valid}} = \cap(L_{AB}, L_{EF}) - \cap(L_{AB}, L_{CD})
\end{equation}
and the corresponding observed angle is:
\begin{equation}
\theta_{\text{valid}} = \text{Angle}_O(L_{\text{valid}})
\end{equation}

\begin{figure}
    \centering
    \includegraphics[width=0.9\linewidth]{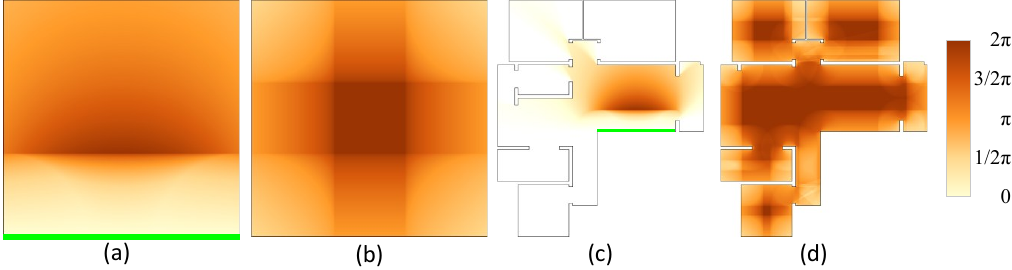}
    \caption{Visibility Field (VF) for static LiDAR scanning. (a, c) VF computed with only one wall (green highlight); (b, d) VF with all walls, revealing the global visibility structure. (a, b) show a rectangular room, and (c, d) depict a flat layout.}
    \label{fig:visibility_field}
\end{figure}

Each viewpoint is assigned a View Angle, with the scene represented as a grid where each cell center is a viewpoint. Thus, the scene is represented as a VF, as shown in \cref{fig:visibility_field}(b), which aligns well with LiDAR properties and expert experience, displaying expected visibility patterns.

\subsection{Converging Lines and Joints}

In the VF, observation data naturally converge toward the medial axis and its joint points. Placing viewpoints along these skeletal structures guarantees full coverage (\cref{sec:theory}, \cref{appendix:theory}), reducing VPN optimization from a 2D search to a 1D problem. We extract the medial axis using \cite{siddiqi2008medial} via OpenCV \cite{bradski2000opencv} and detect joints by restricting inside room polygons: a skeleton point is marked as a joint candidate if at least three of its eight neighbors are also skeleton points. Connected-component analysis groups candidates, and each cluster’s centroid is retained as a representative joint.

For each skeleton ridge \(R_{ij}\) between joints \(J_i\) and \(J_j\), we compute the overlap ratio \(O_{ij}\). If \(O_{ij}\) is below the threshold, a midpoint \(J_m\) is inserted to split the ridge, ensuring connectivity. We refer to all joints and added points as \emph{converging points}, and their connecting ridges as \emph{converging lines}. This process iterates until every converging point satisfies the overlap requirement for reliable registration.

\subsection{Overlap Ratio}\label{sec:overlap}

\begin{figure}[htbp]
    \centering
    \includegraphics[width=0.9\linewidth]{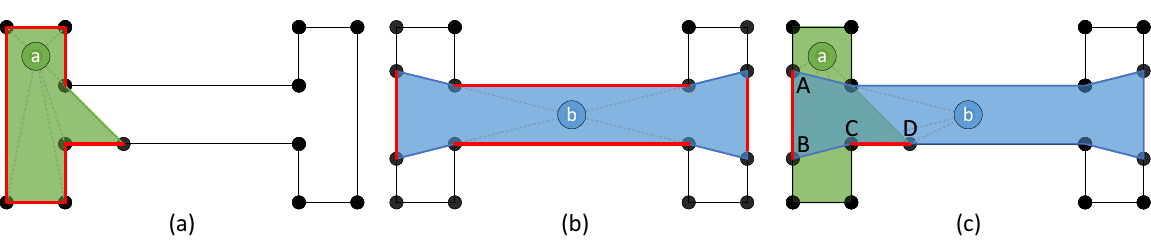}
    \caption{Overlap computation between wall segments from two viewpoints. (a, b) Segments visible from viewpoints A and B. (c) Segments jointly visible, used to measure connectivity.}
    \label{fig:overlap}
\end{figure}

The overlap ratio quantifies shared visibility between two viewpoints and is key for robust registration. We define five variants based on either segment length or angular alignment, evaluated for their ability to maintain connectivity independent of viewpoint spacing.

As shown in \cref{fig:overlap}, $L_a$ and $L_b$ denote wall segments visible from viewpoints $V_a$ and $V_b$, with observation angles $\theta_a$ and $\theta_b$:
\begin{equation}
\theta_a = \text{Angle}(V_a, L_a)
\end{equation}
The segments observed by both \( V_a \) and \( V_b \) are \( L_{ab} \):
\begin{equation}
L_{ab} = L_a \cap L_b
\end{equation}
Length-based ratios:
\begin{equation}
O^{\text{Min\_Len}} = \frac{L_{ab}}{\min(L_a, L_b)}
\end{equation}
\begin{equation}
O^{\text{Mean\_Len}} = \frac{2L_{ab}}{L_a + L_b}
\end{equation}
\begin{equation}
O^{\text{Union\_Len}} = \frac{L_{ab}}{L_a \cup L_b}
\end{equation}
Angle-based ratios, where $\theta_a^{ab}$ and $\theta_b^{ab}$ are the angular spans of $L_{ab}$:
\begin{equation}
O^{\text{Union\_Ang}} = \frac{\theta_a^{ab} + \theta_b^{ab}}{\theta_a + \theta_b}
\end{equation}
\begin{equation}
O^{\text{Mean\_Ang}} = \frac{\theta_a^{ab}}{2\theta_a} + \frac{\theta_b^{ab}}{2\theta_b}
\end{equation}

\subsection{Greedy Optimization for VPN Construction}
The VF-Plan greedy algorithm constructs an efficient VPN by initializing candidate viewpoints at strategically selected joint points within the VF. For each pair of candidate viewpoints, an Overlap Ratio \( O \) is computed to quantify shared visibility. Pairs exceeding a predefined overlap threshold are retained as edges in the VPN, ensuring robust connectivity. To enforce the coverage constraint, we construct a binary Coverage Table \( C \) (see \cref{eq:coverage}) that records whether each viewpoint observes each structural segment.

\begin{figure}[!htbp]
    \centering
    \includegraphics[width=0.9\linewidth]{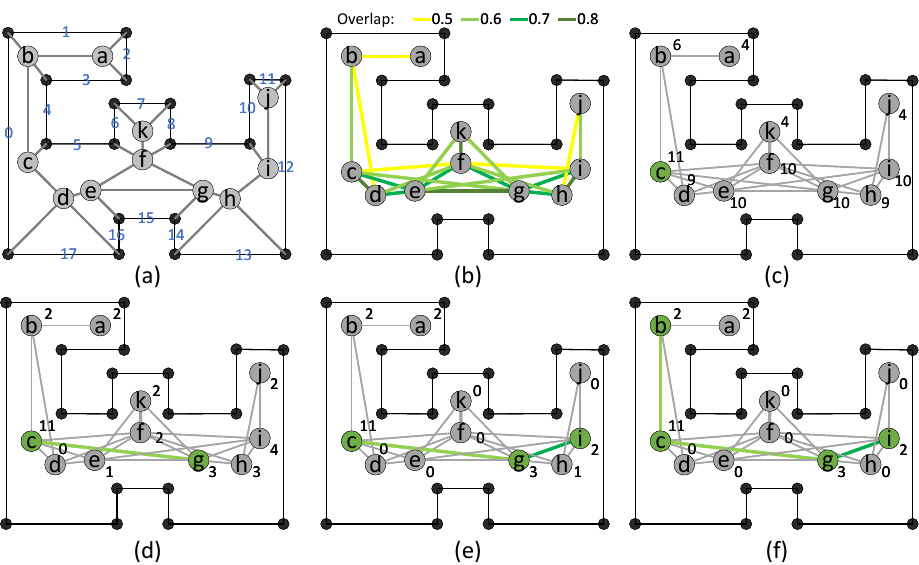}
    \caption{Greedy Optimization of VPN. (a) Floorplan showing candidate viewpoints (labeled a to k) positioned along walls (numbered 0 to 17); (b) Connectivity graph, where edges link viewpoints with overlap ratio \(O\) exceeding the threshold criterion; (c) Initial seed viewpoint selection based on maximum wall visibility (numbers near viewpoints indicate the count of observed walls); (d) - (f) Iterative expansion of the seed viewpoint set by adding adjacent viewpoints that provide the highest additional coverage, enhancing overall viewpoint arrangement.}
    \label{fig:heuristic_search}
\end{figure}

\begin{algorithm} 
\caption{Greedy Optimization of VPN}\label{alg:Heuristic_Optimization}
\textbf{Input:} Candidate viewpoints \( V \), line segments \( L \) \\ 
\textbf{Output:} Minimal set of viewpoints \( V_{\text{opt}} \) covering all \( L \) and forming a connected network

\begin{enumerate}
    \item Initialize Coverage Table \( C \)
    \item Identify \( V_i \) with maximum \( C_i \), add \( V_i \) to Viewpoint Seed Set (VPS)
    \item Add adjacent viewpoints of \( V_i \) to VPB
    \item Remove all \( L \) observed by \( V_i \) from \( C \)
    \item While unobserved \( L \) exists in \( C \):
        \begin{enumerate}
            \item Select \( V_i \) in VPB maximizing new coverage \( C_i \)
            \item If multiple options, select \( V_i \) with highest overlap \( O \) with VPS
            \item Move \( V_i \) to VPS, add adjacencies to VPB
            \item Remove all \( L \) observed by \( V_i \) from \( C \)
        \end{enumerate}
\end{enumerate}

\Return{VPS as the optimized minimal VPN}
\end{algorithm}

As outlined in Algorithm \ref{alg:Heuristic_Optimization} and illustrated in \cref{fig:heuristic_search}, the greedy optimization initiates by selecting the viewpoint with the highest coverage as the seed. The algorithm then iteratively expands by adding neighboring viewpoints that maximize additional coverage, ensuring each new viewpoint significantly contributes to covering line segments while constructing a minimal, fully connected VPN.

Upon completion of the initial search, the algorithm produces a single connected component encompassing all essential viewpoints, with sufficient overlap to facilitate stable connections. The algorithm subsequently enhances network stability by identifying cycles within the VPN graph. For nodes that cannot form cycles (i.e., loose nodes), Converging Points are added to ensure cohesion without redundancy. Additionally, any node with more than three connecting edges from a cycle selects a Converging Joint to establish loops, thereby strengthening network connectivity.

\begin{figure*}[!bthp]
    \centering
    \includegraphics[width=0.8\textwidth]{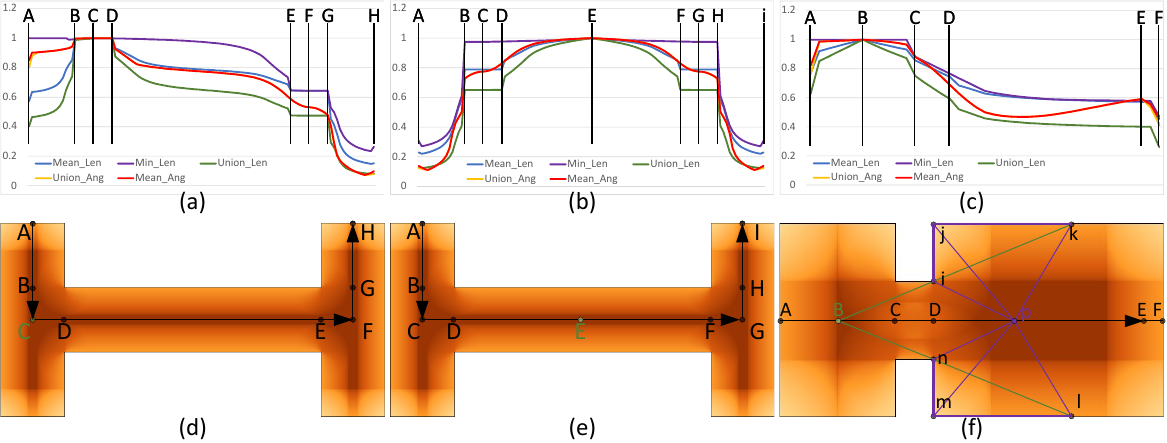}
    \caption{Performance of five proposed overlap ratios. The first row shows overlap ratios between moving and reference viewpoints; the second row illustrates room configurations, with black arrows indicating movement direction and green dots marking reference viewpoints.}
    \label{fig:performance_OR}
\end{figure*}

\section{Experiments}

In this section, we evaluate the proposed VPN optimization algorithm across multiple scenarios and compare its performance with state-of-the-art methods in both indoor and outdoor environments. Due to the lack of a dedicated benchmark for VPN optimization, we compiled a dataset from established sources, focusing on two primary scene types: indoor and outdoor environments. This dataset provides a robust foundation for evaluating VPN optimization across a range of realistic settings. The pipeline is implemented in C++14 and compiled on Windows~11, running on an Intel Core i9 CPU (24 cores, 2.20\,GHz) with 32\,GB RAM.

\subsection{Dataset}
For indoor environments, we selected scenes with varied room layouts to evaluate the algorithm’s adaptability across multiple configurations. The Structure3D dataset~\cite{Structured3D} was included for its extensive collection of diverse Asian residential scenes, featuring both complex and compact room types frequently used in 3D modeling and synthetic data generation. VPNs generated from this dataset can also facilitate synthetic LiDAR data production for advancing research in 3D scene understanding. Specific scenes such as 00007, 00009, 00071, 02970, and 02972, discussed in subsequent sections, are drawn from Structure3D.  For further comparisons, we incorporated additional scenes from Zeng1~\cite{zeng2022optimal}, Noichl~\cite{noichl2023automated}, Xu~\cite{Xu2017}, and TUB1 from the ISPRS benchmark~\cite{zhai2024semantic,ISPRSIndoor}.

For outdoor scenarios, we limited our approach to 2D site plans without considering terrain elevation, distinguishing our method from elevation-based approaches like~\cite{chen20223d}. Selected scenes include the CCIT and Crowsnest datasets from the University of Calgary~\cite{Jia2018} and the ECUT dataset from East China University of Technology~\cite{chen20223d}, which represent open-space environments suitable for VPN optimization.

To evaluate VPN quality, we rely on expert-labeled annotations as standard ground truth data is unavailable. Where possible, we use existing expert-labeled VPNs (e.g., Zeng1, scene1 and TUB1), and for other cases, independent expert labels were created for comparison with manual designs.

\subsection{Performance Metrics}

The optimization algorithm aims to construct a minimal, fully connected VPN with reduced redundancy. As most state-of-the-art methods achieve 100\% coverage, we focus on two key metrics: the number of optimal viewpoints and network connectivity.

Network connectivity is measured using the Weighted Average Path Length (WAPL), which reflects the compactness of the network~\cite{boccaletti2006complex}. A shorter WAPL indicates stronger connectivity and a more efficient structure. For a network with weighted edges \( w \), WAPL is calculated as:
\begin{equation}
WAPL = \frac{1}{N(N - 1)} \sum_{i \neq j} d_w(i, j)
\end{equation}
where \( d_w(i, j) \) represents the shortest weighted path length between nodes \( i \) and \( j \), and \( N \) is the total number of nodes.

The edge weight \( w \) is defined as
\begin{equation}
w_{ij} = 1 - O_{ij}
\end{equation}
A larger \( O_{ij} \) implies stronger registration-based connectivity, meaning a shorter length between the two nodes \( i \) and \( j \).

In cases where baseline methods do not yield a single connected component, some viewpoints may have a small overlap ratio with others. To ensure WAPL remains computable, we assign a large weight (100) to node pairs that lack a connecting path.

\subsection{Ablation Study}

We evaluate overlap ratio type, scan radius, window inclusion, and, in \cref{appendix:more_abltion}, wall partition, VF grid resolution and overlap thresholds. VPN performance is largely insensitive to wall partition length and VF grid resolution, though these parameters affect computation time, while others require tuning for specific scenarios. Default settings for subsequent experiments are: \(O^{\text{Mean\_Len}}\) for overlap computation, overlap ratio \(0.4\) indoors and \(0.3\) outdoors, \(R_{\min}=0.6\)\,m indoors and \(1.2\)\,m outdoors, \(R_{\max}=30.0\)\,m indoors and \(75.0\)\,m outdoors, wall partition \(0.1\)\,m indoors and \(1.0\)\,m outdoors, and VF grid resolution \(0.02\)\,m indoors and \(0.25\)\,m outdoors.

\subsubsection{Overlap Ratio Type}

We propose five overlap ratios according to the length of scanned walls and according angles to viewpoints in \cref{sec:overlap}. Here we evaluate them in two test indoor scenes, illustrated in \cref{fig:performance_OR}. Among them, \( O^{\text{Min\_Len}} \) quickly reaches 100\%, showing limited variability along movement paths, as seen from B to H in \cref{fig:performance_OR}(b) and (e), making it less suitable for dynamic scenes. The angle-based measures, \( O^{\text{Union\_Ang}} \) and \( O^{\text{Mean\_Ang}} \), exhibit non-monotonic behavior, particularly along paths from C to E, shown in \cref{fig:performance_OR}(c) and (f), which limits their usefulness for consistent registration. The \( O^{\text{Union\_Len}} \) measure tends to be too small, often underestimating connectivity. Therefore, we select \( O^{\text{Mean\_Len}} \) for subsequent experiments and analyses due to its balanced performance across different scenarios.

\begin{figure}[bthp]
    \centering
    \includegraphics[width=0.75\linewidth]{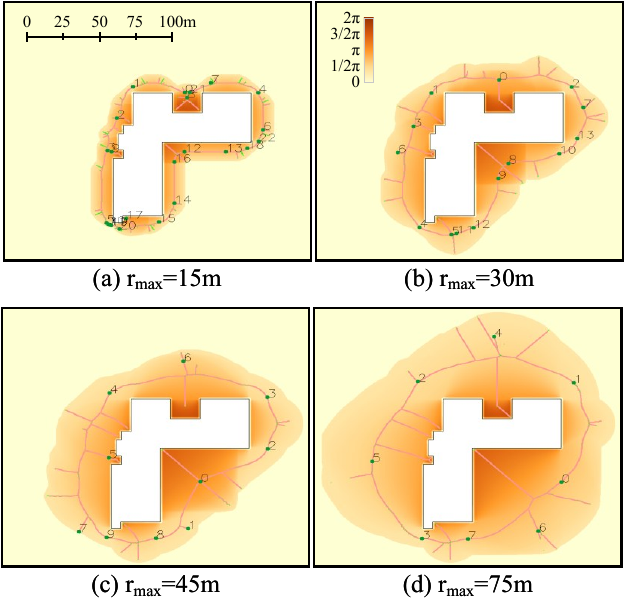}
    \caption{Effect of \( R_{\text{max}} \) on Viewpoint Placement and Network Efficiency. As \( R_{\text{max}} \) increases from 15m to 75m, the valid placement area (where View Angle \(>\) 1.0) grows, reducing the VC needed for optimized VPN coverage from 23 to 8.}
    \label{fig:effect_radius}
\end{figure}

\subsubsection{Scan Radius}
The scan model used in this study operates in the X-Y plane with a 360° field of view, bounded by \( R_{\text{min}} \) and \( R_{\text{max}} \), defining near and far blind zones. Typical static scanners, including the FARO M70, Trimble X7, and Leica BLK360, have maximum ranges around 40 meters and minimum ranges of about 1 meter, allowing for comprehensive indoor coverage but requiring precise placement in small rooms and narrow corridors.

For outdoor scenarios, we examine the impact of varying \( R_{\text{max}} \), testing values at 15m, 30m, 45m, and 75m, with \( R_{\text{min}} \) fixed at 1m. As illustrated in \cref{fig:effect_radius}, increasing \( R_{\text{max}} \) expands the area of valid placements for viewpoints—defined where the View Angle exceeds 1.0—and reduces the Viewpoint Count (VC) needed for optimized coverage, from 23 to 14, 10, and 8. Larger \( R_{\text{max}} \) values are advantageous for open spaces, while smaller values require additional viewpoints for full coverage. These results indicate that our method aligns effectively with real-world scanner specifications and expert guidelines.

\begin{figure}[tbhp]
    \centering
    \includegraphics[width=0.85\linewidth]{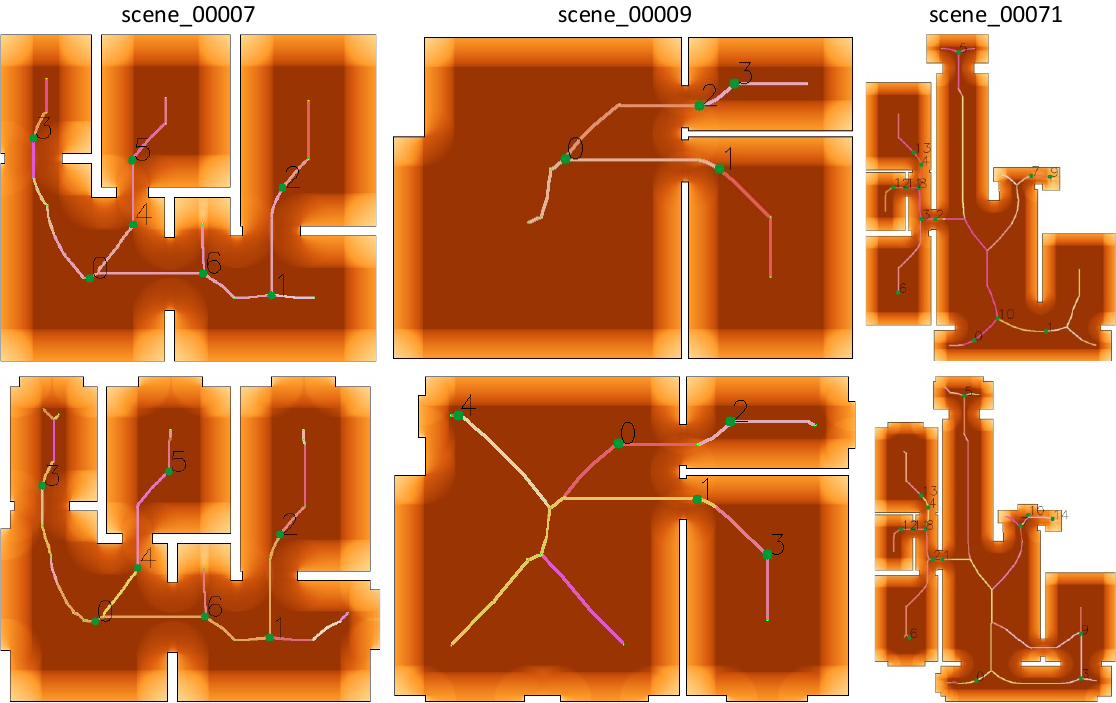}
    \caption{Comparison of VPN structures with and without windows. Including windows (bottom row) adds complexity to the skeleton and slightly increases the VC.}
    \label{fig:VPN_window}
\end{figure}

\subsubsection{Effect of Windows}

Incorporating windows into the model increases the complexity of the skeleton structure by introducing additional joint points, resulting in a more detailed medial axis, as illustrated in \cref{fig:VPN_window}. However, accurate scanning of windows is often essential in practical applications. To assess the impact of including windows, we evaluate the performance of our algorithm with and without windows considered in the optimization. When windows are included, the VPN algorithm requires only a slight increase in the number of viewpoints to achieve complete coverage, as shown in \cref{tab:w_o_window}. On average, the number of viewpoints is approximately 1.4 times the number of rooms without windows, and increases modestly to 1.6 times with windows. These results demonstrate that the algorithm effectively handles the additional geometric detail introduced by windows. For applications requiring precise window coverage, including window frames in the optimization process is recommended.

\begin{table}[!htbp]
    \centering
    \resizebox{0.7\linewidth}{!}{%
    \begin{tabular}{|c|c|c|c|}
        \hline
        Scene & 00007 & 00009 & 00071 \\
        \hline
        Number of Rooms & 5 & 3 & 9 \\
        \hline
        VC (Without Windows) & 7 & 4 & 14 \\
        \hline
        VC (With Windows) & 7& 5 & 15 \\
        \hline
    \end{tabular}%
    }
    \caption{VC comparison for room coverage with and without windows across different scenes.}
    \label{tab:w_o_window}
\end{table}

\subsection{Results and Analysis}

\begin{table}[!htbp]
\centering
\resizebox{\linewidth}{!}{%
\begin{tabular}{|c|c|c|c|c|c|c|c|c|c|c|}
\hline
\multirow{2}{*}{Scene} & \multicolumn{3}{c|}{WAPL$\downarrow$} & \multicolumn{3}{c|}{VC$\downarrow$} & \multirow{2}{*}{\shortstack{Baseline \\ Method}} \\ \cline{2-7}
            & Expert      &Ours   &Baseline     &Expert      &Ours   &Baseline           &           \\ \hline
Zeng1       & 0.47      & 0.53     &43.37         & 12        & 12           &10           & Zeng\cite{zeng2022optimal}    \\ \hline
Noichl      & 0.32      & 0.38     &23.91         &8         & 10            &7           & Noichl2023\cite{noichl2023automated}  \\ \hline
TUB1        & 1.03      &1.00      &40.97         &28         & 29           &19           & Zhai2024\cite{zhai2024semantic}     \\ \hline
Xu          &0.35       & 0.31     &28.69         & 8         & 8            &7           & Xu2017\cite{Xu2017}      \\ \hline
CCIT        &0.52       & 0.47     &38.12         &8          &8             &7           & Jia2018\cite{Jia2018}     \\ \hline
Crowsnest   &0.50       & 0.49     &34.02         & 9         & 8            &8           & Jia2018\cite{Jia2018}     \\ \hline
ECUT        &0.67       & 0.76      &11.94         & 58        & 52          &139           &Chen2023\cite{chen20223d}    \\ \hline
\end{tabular}
}
\caption{Comparison of VC and WAPL across indoor and outdoor scenes. Lower WAPL indicates more compact network connectivity.}
\label{tab:metrics}
\end{table}

We benchmarked our method against state-of-the-art approaches using datasets from prior studies, representing a range of indoor and outdoor scenes. Indoor datasets include Zeng1~\cite{zeng2022optimal}, Noichl~\cite{noichl2023automated}, and TUB1 from the ISPRS dataset~\cite{zhai2024semantic}. Outdoor datasets consist of CCIT and Crowsnest from the University of Calgary~\cite{Jia2018} and ECUT from East China University of Technology~\cite{chen20223d}.

\begin{figure}[tbhp]
    \centering
    \includegraphics[width=1\linewidth]{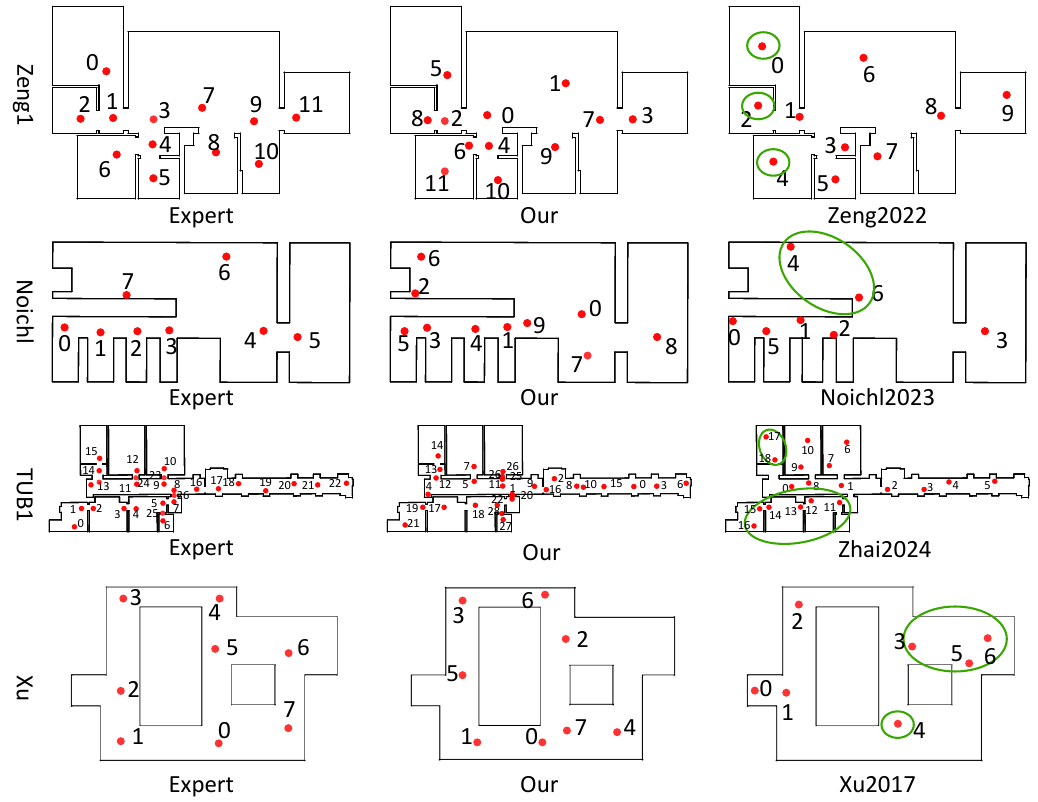}
    \caption{Viewpoint planning results for indoor environments. Red dots represent individual viewpoints, while green ellipses indicate clusters of viewpoints that are not connected to the main network.}
    \label{fig:result_indoor}
\end{figure}

\textbf{Indoor Results.} As shown in \cref{fig:result_indoor}, our method's viewpoint optimization closely aligns with expert-designed networks, achieving comprehensive coverage and creating compact, interconnected structures. This optimized network design enhances registration accuracy and connectivity, even in complex indoor settings.

\begin{figure}[tbhp]
    \centering
    \includegraphics[width=1\linewidth]{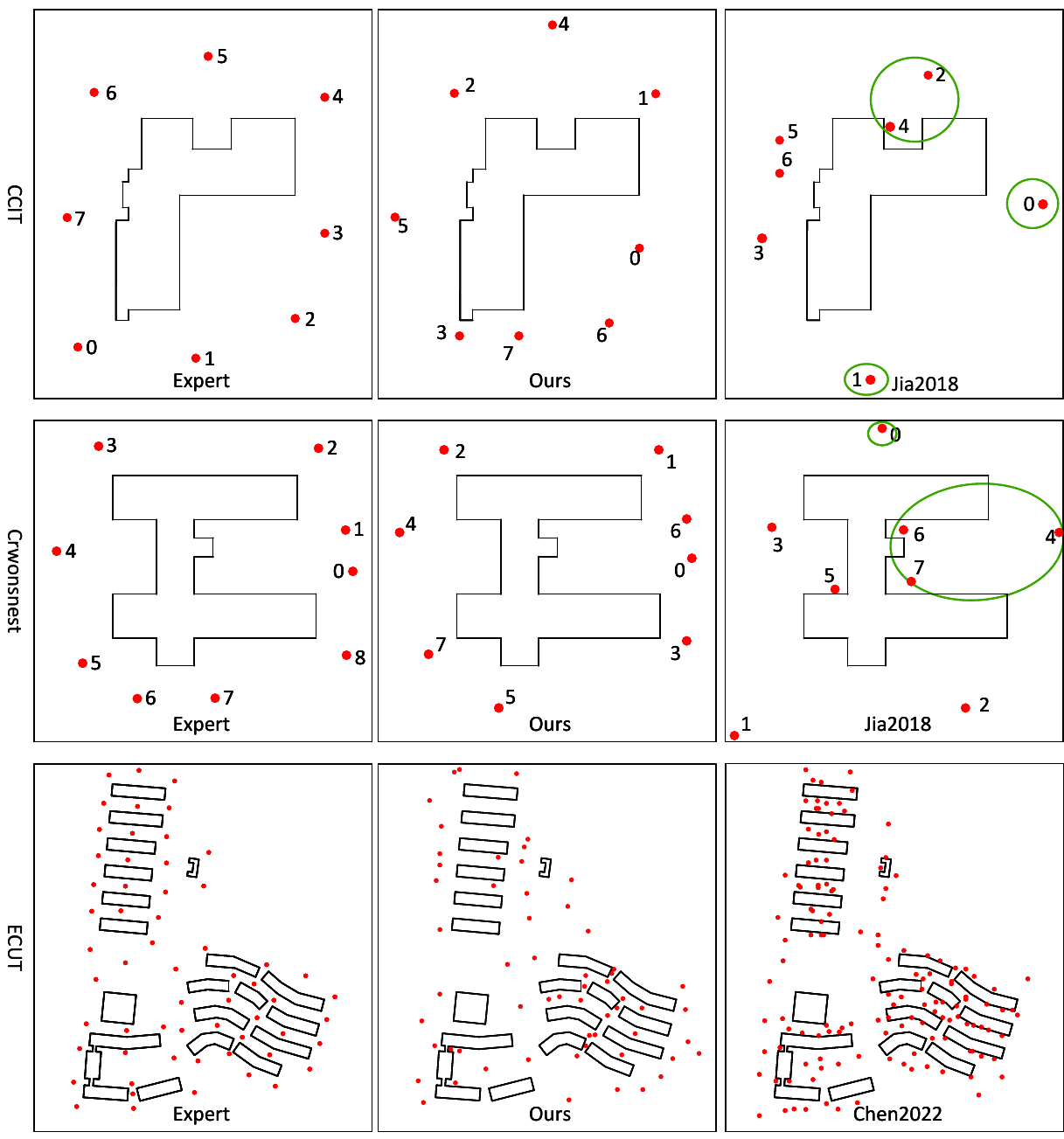}
    \caption{Viewpoint planning results for outdoor environments. Red dots represent individual viewpoints, while green ellipses indicate clusters of viewpoints that are not connected to the main network.}
    \label{fig:result_outdoor}
\end{figure}

\textbf{Outdoor Results.} For outdoor scenes, our method effectively adapts to open spaces, as illustrated in \cref{fig:result_outdoor}. Adjustments to the \( R_{\text{max}} \) parameter notably improved performance in expansive environments, reducing the required number of viewpoints without compromising coverage.

\Cref{tab:metrics} shows that our method matches or surpasses expert-designed networks while significantly outperforming baseline methods. Across all scenes, our VC remains comparable to experts, differing by at most one viewpoint, yet our WAPL is consistently lower in five of seven cases, indicating more compact and cohesive networks. Compared to baselines, the improvement is substantial: in indoor scenes such as Zeng1 and Noichl, WAPL is reduced by 98\% and 97\% respectively, eliminating the fragmented connectivity seen in baseline outputs (green ellipses in \cref{fig:result_indoor,fig:result_outdoor}). In challenging outdoor cases, reductions remain high, including 91\% in ECUT and 87\% in CCIT, demonstrating scalability to large and complex environments. These results confirm that our VF-based optimization preserves expert-level coverage while achieving markedly superior network compactness over existing automated methods.

\begin{figure}[tbhp]
    \centering
    \includegraphics[width=1\linewidth]{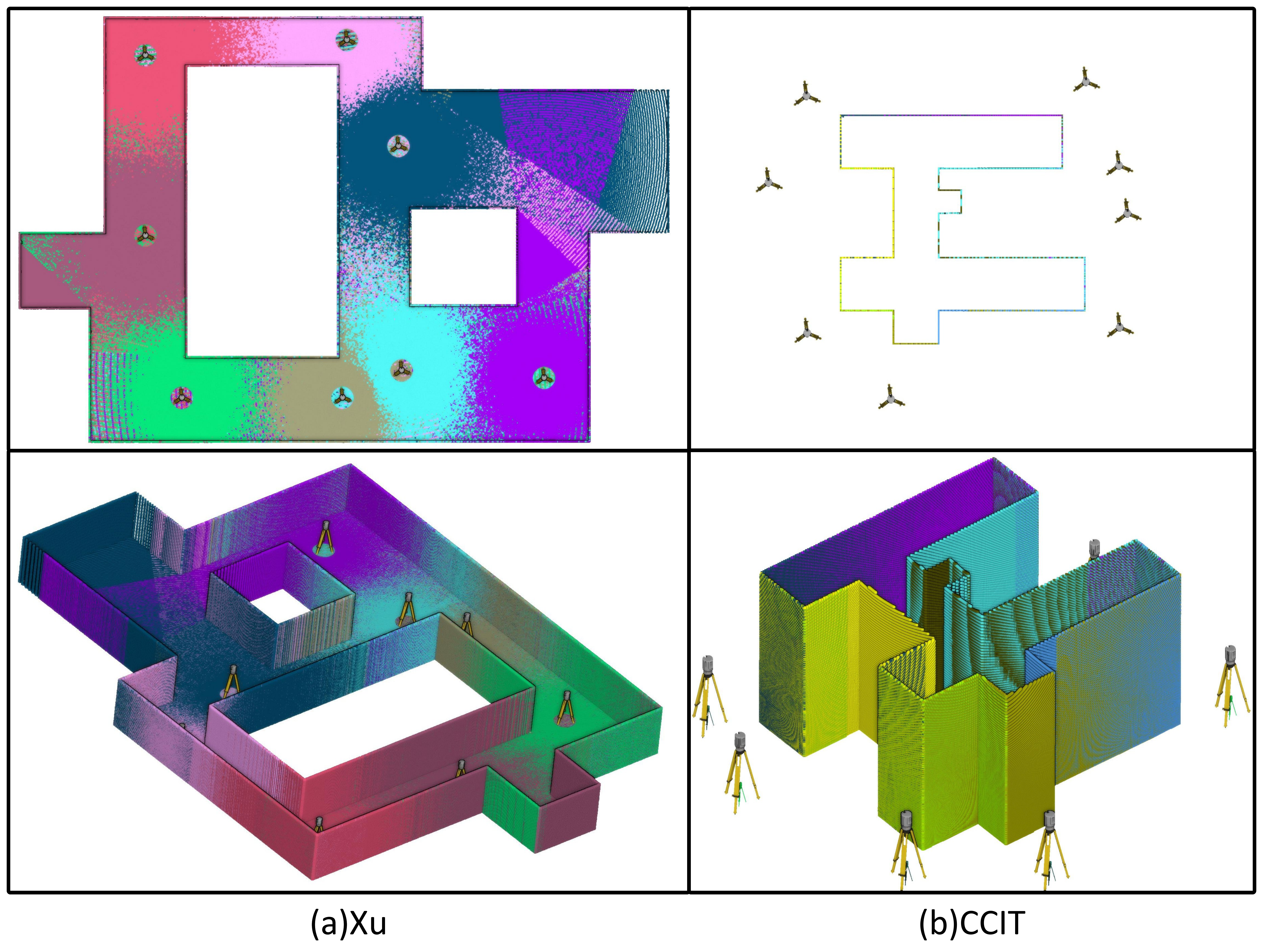}
    \caption{Simulated LiDAR points from optimal viewpoints. Each scan is uniquely colored. Top row shows top views, and bottom row shows oblique views with tripods indicating viewpoints.}
    \label{fig:Simulated}
\end{figure}

To further validate the optimized viewpoints, we simulated static LiDAR scanning at each optimized viewpoint using Helios++~\cite{HELIOS2022}. As shown in \cref{fig:Simulated}, the simulated LiDAR point clouds confirm complete scene coverage, with effective overlap between adjacent scans across both indoor and outdoor environments.

\section{Conclusions}

We presented a novel solution to the VPP and the AGP in static LiDAR scanning on 2D floor plans, leveraging the VF to reduce the optimization complexity from two dimensions to one. This VF-based approach enables the construction of a minimal, fully connected VPN that ensures complete coverage with reduced redundancy, advancing viewpoint planning for static LiDAR applications such as BIM construction. Our method offers a scalable and efficient framework applicable in structured and semi-structured environments, addressing real-world challenges in 3D data collection and autonomous navigation. Our VF framework adapts well to real-world settings with potential obstructions by integrating partial occlusion handling based on initial scans. 

The current formulation assumes accurate and complete 2D floor plans, which may not be available in all scenarios. In cases of noisy input or partial scans, the quality of the VF and the resulting VPN may degrade due to inaccuracies in boundary detection or incomplete visibility information. While preliminary layouts can be generated using low-cost SLAM systems or coarse surveys, further robustness to noise, clutter, and missing geometry remains an open challenge. Additionally, the current implementation focuses on planar (2D) environments; extending the approach to handle 3D structures with significant vertical complexity will require additional algorithmic considerations. Future work will explore its application to large-scale LiDAR data collection and synthetic data generation, as well as integration with path planning in both 2D and 3D scenarios for more versatile deployment in complex environments.

\newpage
\bibliographystyle{eg-alpha-doi}
\bibliography{egbibsample}

\newpage
\setcounter{section}{0}
\def\thesection{\Alph{section}}
\appendix
{\LARGE\textbf{Supplementary Materials}}

\section{Theoretical Foundations, Medial Axis Convergence, and Greedy Optimality}\label{appendix:theory}
The key idea of VF-Plan is that visibility information concentrates on a one dimensional skeleton of the polygonal domain. In the visibility field, visibility propagates from the boundary and accumulates on the medial axis, which reduces the search from two dimensions to one. We formalize this convergence, prove that restricting candidate viewpoints to the medial axis preserves global optimality, and present quantitative performance guarantees for the greedy selection strategy.
\subsection{Skeleton Completeness and Visibility Propagation}\label{sec:skeleton}
\paragraph{Medial Axis (\emph{MA}).}
The medial axis of a polygon \(P\) is the locus of points that are equidistant to at least two boundary elements. In Blum’s grassfire transform~\cite{blum1967} the boundary ignites at time zero and inward fronts meet on the medial axis, which explains the concentration of visibility information on the skeleton.

\begin{theorem}[Skeleton Completeness]\label{thm:extended_skeleton}
Let \(p\in P\) see a set of boundary segments \(L_p\subseteq L\). There exists a finite set \(\{q_\ell\}\subseteq \mathrm{MA}(P)\) such that
\begin{equation}
L_p \subseteq \bigcup_{\ell\in L_p}\mathrm{Vis}(q_\ell).
\end{equation}
Hence any interior coverage provided by \(p\) can be reproduced by viewpoints on the medial axis.
\end{theorem}

\begin{figure*}[tbhp!]
    \centering
    \includegraphics[width=0.9\linewidth]{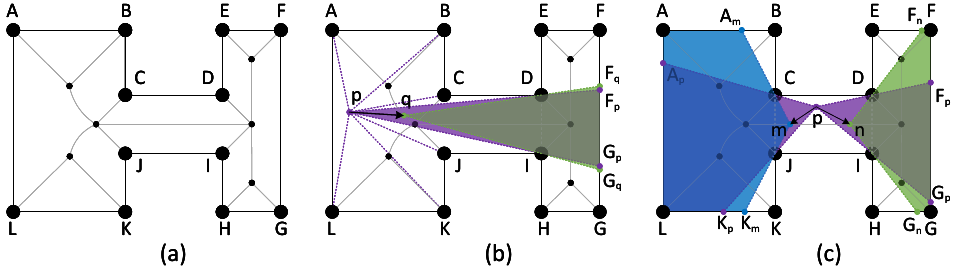}
    \caption{Illustration of skeleton completeness. (a) A simple polygon (black boundary) with its medial axis (gray arcs). (b) Coverage by an interior viewpoint \(p\) can be replicated by a single skeletal viewpoint \(q\) lying on the medial axis (MA). (c) In the general case, coverage from an interior viewpoint \(p\) may require multiple skeletal viewpoints (here, \(m\) and \(n\)) to fully reproduce \(p\)'s visibility. Thus, no coverage capability is lost by restricting candidate viewpoints to the MA. For clarity, boundary segments commonly observed by both \(p\) and the MA viewpoints (\(q\), or \(m\) and \(n\)) are not drawn.}
    \label{fig:proof_skeleton}
\end{figure*}

\noindent\textbf{Sketch of Proof.}
By the grassfire view~\cite{blum1967} visibility from each boundary segment propagates inward until it meets \(\mathrm{MA}(P)\). For every \(l_j\in L_p\) there exists \(q_j\in \mathrm{MA}(P)\) that sees \(l_j\). Therefore \(\mathrm{Vis}(p)\subseteq \bigcup_{l_j\in \mathrm{Vis}(p)}\mathrm{Vis}(q_j)\). Figure~\ref{fig:proof_skeleton} illustrates the construction. Restricting candidates to \(\mathrm{MA}(P)\) does not reduce achievable coverage.

\subsection{Global Minimal Cover and Connectivity on the Skeleton}\label{sec:global_optimality}

By \cref{thm:extended_skeleton} any feasible interior solution transfers to the medial axis. In the \cref{eq:problem_formulation} restricting \(V\) to \(\mathrm{MA}(P)\) including branch and joint nodes therefore captures all feasible solutions.

\begin{theorem}[Global Minimal Cover and Connectivity]\label{thm:global_opt}
Let \(S^*\subseteq \mathrm{MA}(P)\) be a minimal solution to \cref{eq:problem_formulation} that covers all \(l_j\in L\) and satisfies the required overlap–based connectivity. Then \(S^*\) is irreducible and no smaller configuration in \(P\) can achieve the same objectives. Therefore \(S^*\) is globally optimal.
\end{theorem}

\paragraph{Implication and heuristic search.}
No point outside \(\mathrm{MA}(P)\) improves coverage or connectivity over a minimal skeletal solution. Solving \cref{eq:problem_formulation} on \(\mathrm{MA}(P)\) yields a configuration that is optimal for the whole polygon. The problem remains NP–hard, yet a greedy strategy is effective in practice and admits quantitative guarantees for the number of viewpoints.

\subsection{Greedy Selection and Suboptimality Bounds}\label{sec:greedy_bounds}

Let \(n=|L|\) be the number of discretized boundary segments. Ignoring connectivity, selecting viewpoints to cover \(L\) is a set cover instance. The classic greedy rule achieves an approximation factor \(H_n\le 1+\ln n\) for the number of selected sets~\cite{chvatal1979greedy}. Our solver first runs greedy coverage on medial–axis candidates and then augments connectivity on the induced viewpoint graph whose edge weights are \(w_{ij}=1-O_{ij}\) where \(O_{ij}\) is the overlap ratio. Connecting the greedy terminals with a Steiner tree in this graph adds only connector viewpoints along medial–axis paths. Using a constant–factor approximation for Steiner tree on graphs, the number of added connectors is within a constant factor of the optimum connector count. Combining the two stages gives
\begin{equation}
|S_{\text{greedy}}| \;\le\; H_n\cdot \mathrm{OPT}_{\text{cover}} \;+\; \alpha\,\mathrm{OPT}_{\text{conn}},
\end{equation}
where \(\alpha\) is a constant and \(\mathrm{OPT}_{\text{cover}}\) and \(\mathrm{OPT}_{\text{conn}}\) are the minimal numbers of coverage and connector viewpoints on \(\mathrm{MA}(P)\). When connectivity is not stricter than coverage, \(\mathrm{OPT}_{\text{conn}}\le \mathrm{OPT}_{\text{cover}}\), which yields the compact bound
\begin{equation}
|S_{\text{greedy}}| \;\le\; (H_n+\alpha)\,\mathrm{OPT},
\end{equation}
with \(\mathrm{OPT}\) the minimal skeletal solution of \cref{eq:problem_formulation}. Empirically our solutions match or improve expert VC in five of seven scenes and reduce WAPL by up to \(98\%\) relative to automated baselines while remaining within one viewpoint of expert designs in most cases, which corroborates the tightness of the bound in practice.

\section{More Ablation Study}\label{appendix:more_abltion}

\subsubsection{Effect of Wall Partition}

\begin{figure}[!hbt]
    \centering
    \includegraphics[width=1\linewidth]{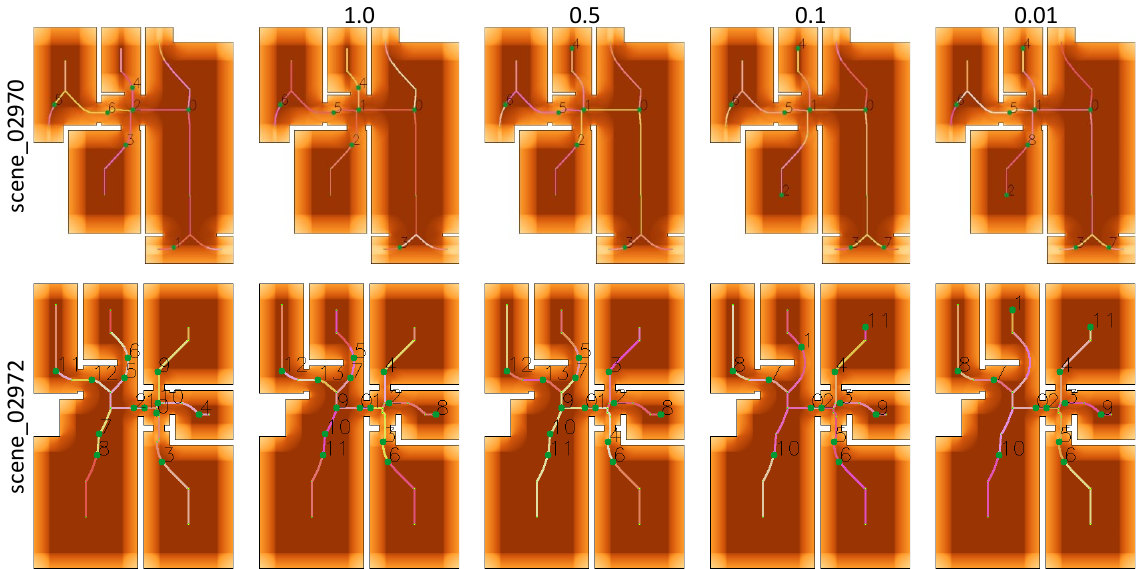}
    \caption{Effects of wall partitions. From left to right: unpartitioned walls and walls partitioned into segments of 1.0m, 0.5m, 0.1m, and 0.01m.}
    \label{fig:wall_partitions}
\end{figure}

To evaluate the impact of wall partition length, wall boundaries are divided into segments of varying lengths including 1.0\,m, 0.5\,m, 0.1\,m, and 0.01\,m, as illustrated in \cref{fig:wall_partitions}. Experimental results show that the partition length does not affect the final optimization outcome, due to the continuous nature of the VF employed in our method. The resolution of partitioning influences only the order in which viewpoints are selected. Coarser partitions, corresponding to longer segments or no partitioning, tend to place initial viewpoints at major structural intersections. A coarse partition reduces computational cost while maintaining effective coverage. Based on this observation, we set the default partition length to 0.1\,m for indoor scenes and 1.0\,m for outdoor scenes in all subsequent experiments.

\subsection{Effect of Overlap Ratio Threshold}\label{appendix:ablation_or}

\begin{figure*}[t]
    \centering
    \includegraphics[width=1\linewidth]{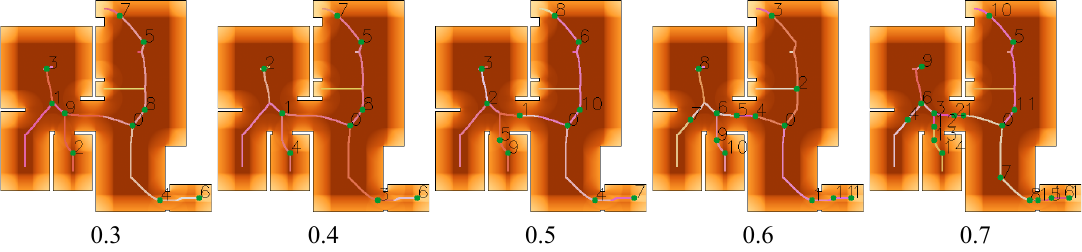}
    \caption{Effect of overlap ratio (OR) on viewpoint planning in Scene\_00001. Increasing OR from 0.3 to 0.7 improves connectivity but raises viewpoint count (VC), showing a trade-off between coverage and compactness, while WAPL and runtime remain stable.}
    \label{fig:ablation_or}
\end{figure*}

We analyze the influence of overlap ratio (OR) on VPN optimization using the indoor Scene\_00001 dataset. The OR defines the minimum visible segment proportion shared between two viewpoints for them to be considered connected. Experiments were conducted for OR values ranging from $0.3$ to $0.7$, with all other parameters fixed: $R_{\min}=0.6$\,m, $R_{\max}=30.0$\,m, resolution $0.02$\,m, and wall partition length $0.2$\,m. Results in \cref{tab:abl_overlapratio,fig:overlap_bars,fig:ablation_or} show that increasing OR generally enhances network connectivity but often requires more viewpoints to maintain full coverage. For example, raising OR from $0.3$ to $0.7$ increases the number of viewpoints from $10$ to $18$ (an 80\% increase) while WAPL rises from $0.38$ to $0.44$. Moderate OR values, such as $0.4$, achieve a balanced trade-off with only $9$ viewpoints, a WAPL of $0.40$, and stable runtime ($\approx0.74$\,s across all settings). Very high OR values ($\geq0.6$) lead to sharp viewpoint growth without proportional gains in WAPL, indicating diminishing returns in connectivity improvement. These observations suggest that an OR in the range $0.3$–$0.5$ is sufficient for most scanning tasks, providing robust connectivity while keeping network size compact.

\begin{table}[thbp]
\centering
\small
\begin{tabular}{cccccc}
\toprule
\textbf{OR} & 0.3 & 0.4 & 0.5 & 0.6 & 0.7 \\
\midrule
VC & 10 & 9  & 11 & 13 & 18 \\
WAPL      & 0.38 & 0.40 & 0.45 & 0.45 & 0.44 \\
Time (s)  & 0.72& 0.75& 0.73& 0.74& 0.74\\
\bottomrule
\end{tabular}
\caption{Effect of overlap ratio (OR) on Scene\_00001.}
\label{tab:abl_overlapratio}
\end{table}


\definecolor{nm-blue}{HTML}{0072B2}   
\definecolor{nm-orange}{HTML}{E69F00} 
\definecolor{nm-green}{HTML}{009E73}  

\begin{figure}[thbp]
\centering
\begin{tikzpicture}
\begin{axis}[
    font=\footnotesize,
    ybar,
    bar width=10pt,
    width=1\linewidth,      
    height=6.2cm,
    ymin=0, ymax=20,       
    symbolic x coords={0.3,0.4,0.5,0.6,0.7},
    xtick=data,
    xticklabel style={/pgf/number format/fixed},
    enlarge x limits=0.12,
    axis on top,
    xlabel={OR},
    xlabel style={
        at={(axis description cs:1,0)}, 
        anchor=west,
        yshift=2ex                   
    },
    tick style={black},
    major tick length=2pt,
    ytick pos=left,
    xtick pos=bottom,
    nodes near coords,
    legend columns=3,
    legend style={
        at={(0.5,1.03)}, anchor=south,
        draw=none, fill=none,
        /tikz/column sep=6pt,  
        font=\footnotesize
    },
    legend cell align={left},
    legend image code/.code={\draw[#1,fill=#1] (0cm,-0.1cm) rectangle (0.18cm,0.18cm);}
]

\addplot[fill=nm-blue,  draw=nm-blue]  coordinates {(0.3,10) (0.4,9) (0.5,11) (0.6,13) (0.7,18)};

\addplot[fill=nm-orange,draw=nm-orange] coordinates {(0.3,3.8) (0.4,4.0) (0.5,4.5) (0.6,4.5) (0.7,4.4)};

\addplot[fill=nm-green, draw=nm-green] coordinates {(0.3,7.2) (0.4,7.5) (0.5,7.3) (0.6,7.4) (0.7,7.4)};

\legend{VC, 10$\times$WAPL, 10$\times$Time (s)}
\end{axis}
\end{tikzpicture}

\caption{Effect of overlap ratio (OR) on Scene\_00001. WAPL and runtime are multiplied by 10 for scale comparability.}

\label{fig:overlap_bars}
\end{figure}

\subsection{Effect of VF Grid Resolution}\label{appendix:grid_resolution}
\begin{figure*}[tbhp]
    \centering
    \includegraphics[width=1\linewidth]{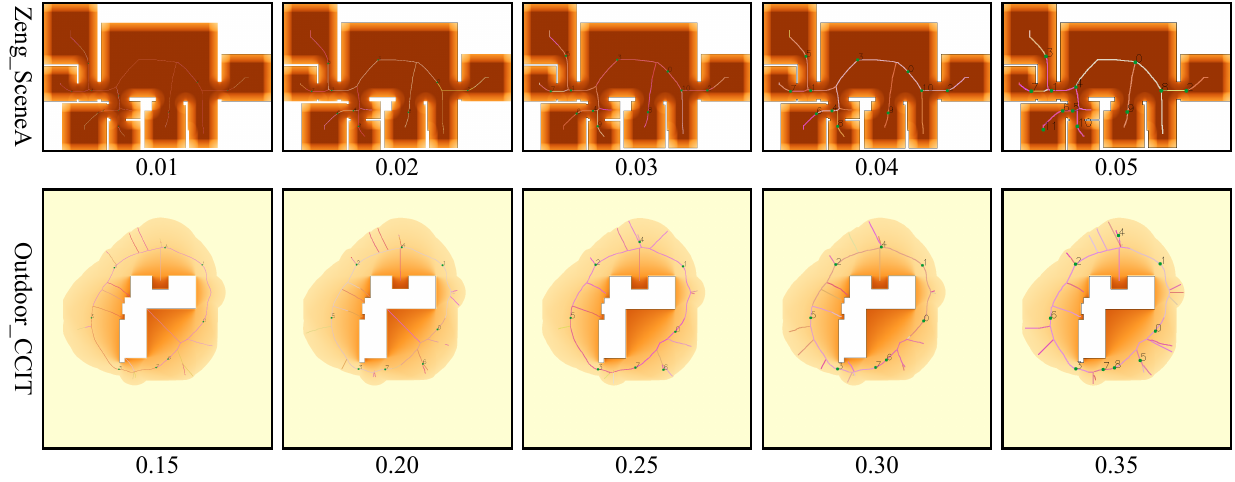}
    \caption{Effect of resolution on VF-based viewpoint planning. The top row shows results for the indoor Zeng\_SceneA case with resolutions ranging from 0.01\,m to 0.05\,m, and the bottom row shows results for the outdoor CCIT case with resolutions ranging from 0.15\,m to 0.35\,m. The VF’s topological nature results in low sensitivity to resolution changes in both scenarios.}
    \label{fig:resolution_effect_vis}
\end{figure*}

We examine the impact of VF grid resolution on VPN optimization using the indoor Zeng\_SceneA and outdoor CCIT datasets. For Zeng\_SceneA, parameters are $R_{\min}=0.6$\,m, $R_{\max}=30.0$\,m, overlap ratio $0.4$, and wall partition $0.2$\,m. For CCIT, parameters are $R_{\min}=1.2$\,m, $R_{\max}=75.0$\,m, overlap ratio $0.3$, and wall partition $1.0$\,m. All other settings remain fixed, and only the grid resolution is varied. As shown in \cref{tab:resolution_effect,fig:resolution_bars_Zeng,fig:resolution_bars_CCIT,fig:resolution_effect_vis}, resolution changes have minimal influence on VPN count or WAPL, with variations under $5\%$ for both datasets. In Zeng\_SceneA, VPN count ranges from $10$ to $12$ and WAPL from $0.52$ to $0.46$ as resolution decreases from $0.01$\,m to $0.05$\,m, while runtime drops from $1.57$\,min to $0.03$\,min. In CCIT, VPN count varies from $8$ to $9$ and WAPL from $0.43$ to $0.44$ as resolution decreases from $0.15$\,m to $0.35$\,m, with runtime falling from $0.21$\,min to $0.03$\,min. Runtime scales approximately quadratically with grid resolution due to the growth in visibility evaluations. These results confirm that the VF-based approach depends primarily on the topological structure of the floorplan, allowing the use of lower resolutions that capture necessary geometric detail while significantly improving runtime.

 \begin{table*}[htbp]
\centering
\caption{Effect of resolution on Viewpoint count, WAPL, and runtime for indoor (Zeng\_SceneA) and outdoor (Outdoor\_CCIT) cases.}
\begin{tabular}{c|ccccc|ccccc}
\toprule
 & \multicolumn{5}{c|}{Zeng\_SceneA} & \multicolumn{5}{c}{Outdoor\_CCIT} \\
\midrule
Resolution (m) & 0.01 & 0.02 & 0.03 & 0.04 & 0.05 & 0.15 & 0.20 & 0.25 & 0.30 & 0.35 \\
\midrule
VC & 11 & 10 & 11 & 11 & 12 & 8 & 8 & 8 & 8 & 9 \\
WAPL & 0.52 & 0.52 & 0.49 & 0.48 & 0.46 & 0.43 & 0.45 & 0.43 & 0.43 & 0.44 \\
Time (min) & 1.57& 0.16& 0.07& 0.04& 0.03& 0.21& 0.08& 0.05& 0.03& 0.03\\
\bottomrule
\end{tabular}
\label{tab:resolution_effect}
\end{table*}

\definecolor{nm-blue}{HTML}{0072B2}   
\definecolor{nm-orange}{HTML}{E69F00} 
\definecolor{nm-green}{HTML}{009E73}  

\begin{figure}[thbp]

\begin{tikzpicture}
\begin{axis}[
    font=\footnotesize,
    ybar,
    bar width=9pt,
    width=1\linewidth,
    height=6.2cm,
    ymin=0, ymax=18,
    xlabel={(m)},
    xlabel style={
        at={(axis description cs:1,0)}, 
        anchor=west,
        yshift=2ex 
    },
    symbolic x coords={0.01,0.02,0.03,0.04,0.05},
    xtick=data,
    xticklabel style={/pgf/number format/fixed},
    enlarge x limits=0.12,
    axis on top,
    tick style={black},
    major tick length=2pt,
    ytick pos=left,
    xtick pos=bottom,
    nodes near coords,
    legend columns=3,
    legend style={
        at={(0.0,1.05)},
        anchor=west,
        draw=none,
        fill=none,
        font=\footnotesize,
        /tikz/column sep=3pt
    },
    legend cell align={left},
    legend image code/.code={
        \draw[#1,fill=#1] (0cm,-0.1cm) rectangle (0.15cm,0.15cm);
    }
]
\addplot[fill=nm-blue,draw=nm-blue] coordinates
{(0.01,11) (0.02,10) (0.03,11) (0.04,11) (0.05,12)};
\addplot[fill=nm-orange,draw=nm-orange] coordinates
{(0.01,5.2) (0.02,5.2) (0.03,4.9) (0.04,4.8) (0.05,4.6)};
\addplot[fill=nm-green,draw=nm-green] coordinates
{(0.01,15.7) (0.02,1.6) (0.03,0.7) (0.04,0.4) (0.05,0.3)};
\legend{VC, 10$\times$WAPL, 10$\times$Time (min)}
\end{axis}
\end{tikzpicture}

\caption{Effect of resolution on Viewpoint count (VC), WAPL, and runtime for indoor (Zeng\_SceneA).}
\label{fig:resolution_bars_Zeng}
\end{figure}
\definecolor{nm-blue}{HTML}{0072B2}   
\definecolor{nm-orange}{HTML}{E69F00} 
\definecolor{nm-green}{HTML}{009E73}  

\begin{figure}[thbp]
\centering
\begin{tikzpicture}
\begin{axis}[
    font=\footnotesize,
    ybar,
    bar width=9pt,
    width=1\linewidth,
    height=6.2cm,
    ymin=0, ymax=10,
    xlabel={(m)},
    xlabel style={
        at={(axis description cs:1,0)}, 
        anchor=west,
        yshift=2ex                   
    },
    symbolic x coords={0.15,0.20,0.25,0.30,0.35},
    xtick=data,
    xticklabel style={/pgf/number format/fixed},
    enlarge x limits=0.12,
    axis on top,
    tick style={black},
    major tick length=2pt,
    ytick pos=left,
    xtick pos=bottom,
    nodes near coords,
    legend columns=3,
    legend style={
        at={(0.0,1.05)},
        anchor=west,
        draw=none,
        fill=none,
        font=\footnotesize,
        /tikz/column sep=3pt
    },
    legend cell align={left},
    legend image code/.code={
        \draw[#1,fill=#1] (0cm,-0.1cm) rectangle (0.15cm,0.15cm);
    }
]
\addplot[fill=nm-blue,draw=nm-blue] coordinates
{(0.15,8) (0.20,8) (0.25,8) (0.30,8) (0.35,9)};
\addplot[fill=nm-orange,draw=nm-orange] coordinates
{(0.15,4.3) (0.20,4.5) (0.25,4.3) (0.30,4.3) (0.35,4.4)};
\addplot[fill=nm-green,draw=nm-green] coordinates
{(0.15,2.1) (0.20,0.8) (0.25,0.5) (0.30,0.3) (0.35,0.3)};
\legend{VC, 10$\times$WAPL, 10$\times$Time (min)}
\end{axis}
\end{tikzpicture}

\caption{Effect of resolution on Viewpoint count (VC), WAPL, and runtime for outdoor (CCIT).}
\label{fig:resolution_bars_CCIT}
\end{figure}

\subsection{Time and Space Complexity Analysis}\label{appendix:complex}

\begin{figure*}[tbhp]
    \centering
    \includegraphics[width=1\linewidth]{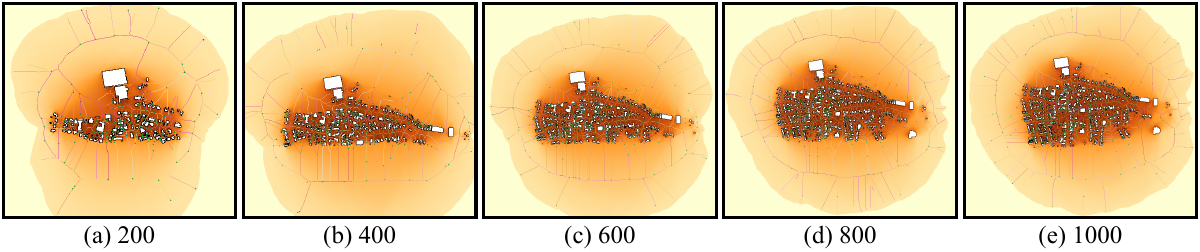}
    \caption{Optimized VPNs for the town dataset. Panels (a) to (e) show cropped regions for viewpoint counts \( \mathrm{VC}=200, 400, 600, 800, 1000 \) respectively.}
    \label{fig:resolution_effect_rst}
\end{figure*}

We evaluate the computational complexity of VF-based VPN optimization in a large-scale outdoor town scenario. To assess scalability, the original dataset is cropped into subregions with approximately \(200, 400, 600, 800,\) and \(1000\) viewpoints (VC). The full town comprises 637 buildings within a block of \(976\,\text{m}\times893\,\text{m}\), totaling \(80{,}549\,\text{m}^2\) of built area with an average footprint of \(126\,\text{m}^2\). All parameters are fixed across experiments: VF resolution \(2.0\,\text{m}\), minimum scan radius \(0.6\,\text{m}\), maximum scan radius \(600\,\text{m}\), wall density \(5\,\text{m}\), and overlap ratio \(0.3\). This setup enables controlled scaling of VC while preserving realistic geometric complexity (\cref{tab:resolution_effect,fig:resolution_effect,fig:resolution_effect_rst}).

\Cref{tab:resolution_effect} shows that memory usage grows nearly linearly with VC, from \(265\,\text{MB}\) at VC=200 to \(5703\,\text{MB}\) at VC=1000 (a 21.5× increase). On a logarithmic scale, \(\log_{10}(\text{RAM})\) increases by only 1.34 across the five settings, confirming mild scaling. By contrast, runtime grows superlinearly: from \(2.68\,\text{min}\) at VC=200 to \(1505.36\,\text{min}\) (\(\approx25\,\text{hrs}\)) at VC=1000, corresponding to a 561× increase. The logarithmic slope of \(\log_{10}(\text{Runtime})\) averages 0.55 per VC-doubling, consistent with near-quadratic growth due to pairwise visibility evaluations.

Comparative scaling further highlights runtime as the primary bottleneck. For example, increasing VC from 400 to 800 multiplies memory by 3.2× (1018 to 3268 MB) but runtime by 31× (30.8 to 967.5 min). Despite this, memory remains within typical workstation limits (<6 GB), while runtime quickly becomes prohibitive at VC \(\geq 1000\). These findings demonstrate that VF-based optimization is tractable at city-block scales but requires algorithmic acceleration—such as hierarchical visibility pruning, parallel BSP-tree queries, GPU acceleration, or distributed processing—for deployment in very large regions.

\begin{figure}[t]
\centering
\begin{tikzpicture}
\begin{axis}[
    font=\footnotesize,
    width=\linewidth,
    height=6.0cm,
    xlabel={VC},
    xlabel style={
        at={(axis description cs:1.05,0)}, 
        anchor=west,
        yshift=2.2ex                   
    },
    xmin=180, xmax=1020,
    ymin=0,   ymax=4.0,
    xtick={200,400,600,800,1000},
    ymajorgrids=true,
    grid style={gray!20},
    tick style={black},
    legend style={
        at={(0.98,0.02)}, 
        anchor=south east,
        draw=none, fill=none
    }
]
\addplot[semithick, nm-blue, mark=*, mark size=2.2pt]
coordinates {(200,3.01) (400,3.23) (600,3.43) (800,3.51) (1000,3.76)};
\addlegendentry{$\log_{10}\!\big(\mathrm{RAM}/\mathrm{MB}\big)$}

\addplot[semithick, nm-orange, mark=square*, mark size=2.2pt]
coordinates {(200,0.43) (400,1.49) (600,1.85) (800,2.99) (1000,3.18)};
\addlegendentry{$\log_{10}\!\big(\mathrm{Runtime}/\mathrm{min}\big)$}
\end{axis}
\end{tikzpicture}
\caption{Log-scaled memory and runtime vs. viewpoint count (VC) for the town dataset. Both $\log_{10}$(RAM/MB) and $\log_{10}$(Runtime/min) increase with CV, highlighting time and space complexity trends.}
\label{fig:resolution_effect}
\end{figure}

\begin{table*}[htbp]
\centering
\caption{Time and space complexity analysis for the town dataset with cropped regions.}
\label{tab:resolution_effect}
\begin{tabular}{r r r c r c c c c c}
\toprule
VC & \makecell{VC\\(Actual)} & \makecell{RAM\\$[MB]$}  &
\makecell{$\log_{10}$\\(RAM/MB)} &
\makecell{Runtime\\ $[min]$} &
\makecell{$\log_{10}$\\(Runtime/min)} &
\makecell{Building\\Count} & \makecell{Block Area\\$[m \times m]$} & \makecell{Total Building\\Area $[m^2]$} &\makecell{Avg Area \\Building $[m^2]$} \\
\midrule
200  &197 & 265 & 2.42 &    2.68 & 0.43 & 132 & $520\times292$ & 22059 & 167 \\
400  &409 &1018 & 3.01 &   30.84 & 1.49 & 274 & $965\times366$ & 39359 & 144 \\
600  &553 &1683 & 3.23 &   71.02 & 1.85 & 379 & $976\times441$ & 49428 & 130 \\
800  &757 &3268 & 3.51 &  967.48 & 2.99 & 529 & $976\times588$ & 67448 & 128 \\
1000 &954 &5703 & 3.76 & 1505.36 & 3.18 & 637 & $976\times893$ & 80549 & 126 \\
\bottomrule
\end{tabular}%
\end{table*}

\end{document}